\documentclass[a4paper]{article}
\usepackage[english]{babel}
\usepackage[T1]{fontenc}

\usepackage[a4paper,top=3cm,bottom=2cm,left=3cm,right=3cm,marginparwidth=1.75cm]{geometry}

\usepackage{amsmath}
\usepackage{graphicx}
\usepackage[colorinlistoftodos]{todonotes}
\usepackage[colorlinks=true, allcolors=blue]{hyperref}
\title{Reproduction Report on "Learn To Pay Attention"}

\author{
  Levan Shugliashvili, Davit Soselia, Shota Amashukeli, Irakli Koberidze \\
  SDSU Georgia Artificial Intelligence and Machine Learning Club\\
  Tbilisi, Georgia \\
  \texttt{\{lshugliashvili4659, dsoselia, samashukeli, ikoberidze\}@sdsu.edu }\\
}
\begin{document}
\maketitle

\begin{abstract}
We have successfully implemented the "Learn to Pay Attention" \cite{src} model of attention mechanism in convolutional neural networks, and have replicated the results of the original paper in the categories of image classification and fine-grained recognition. 
\end{abstract}

\section{Introduction}

The model proposed in the "Learn to Pay Attention" paper introduced a novel way to generate a trainable attention module for convolutional neural networks. The paper demonstrated the attention module in VGG-based and ResNet-based architectures, provided several options for implementation (including three options for layer depths at which attention modules are to be implemented, two options for calculating the compatibility between the global and local features in generating the attention maps, and two options for what method will be used to produce output probabilities from global-level feature vectors), described the dataset preprocessing and model training routines, and reported the results of the consequent models in several tasks. We have successfully implemented all possible configurations of both VGG-based and ResNet-based attention models, and have replicated the paper's reported results in image classification and fine-grain recognition task using the (VGG-att2)-concat-pc configuration on the CIFAR-10 dataset and (VGG-att3)-concat-pc configuration on CIFAR-100 and the SVHN dataset. The code for our reproduction is available at \hyperlink{https://github.com/DadianisBidza/LearnToPayAttention-Keras}{https://github.com/DadianisBidza/LearnToPayAttention-Keras}.

\section{Implementation}

\subsection{Keras functional model}

We used Keras functional API to implement the Learn To Pay Attention model, with a custom ParametrisedCompatibility layer for implementing the parametrized compatibility scheme described in the paper and a custom LearningRateScaler callback for implementing the learning rate schedule described in the paper. Although the paper described using the initial learning rate of 1 for CIFAR and 0.1 for SVHN, we found that our implementation did not converge unless we used much lower initial learning rates of 0.01 for CIFAR and 0.0025 for SVHN. (we attribute this to difference in implementation, as at lower initial learning rates our model's performance matched the performance described in the paper). Table 1 describes the hyperparameters of the Learn to Pay Attention model.

\begin{table}[h]
\begin{tabular}{lllll}
\cline{1-2}
\multicolumn{1}{|l|}{att}           & \multicolumn{1}{l|}{\begin{tabular}[c]{@{}l@{}}The number of levels at which the attention \\ module will be attached. Possible values: att1, att2, att3.\end{tabular}}                                                                                                                                                         &  &  &  \\ \cline{1-2}
\multicolumn{1}{|l|}{compatibility\_function} & \multicolumn{1}{l|}{\begin{tabular}[c]{@{}l@{}}The method that will be used for calculating between \\ local and global features. \\ Possible values: dp (using dot product between \\ the global and the local features),\\ pc (using the paper's described  parametrised compatibility process).\end{tabular}}                      &  &  &  \\ \cline{1-2}
\multicolumn{1}{|l|}{g\_mode}       & \multicolumn{1}{l|}{\begin{tabular}[c]{@{}l@{}}Method for using the local features to generate prediction. \\ Possible values:  concat (local features are concatenated and \\ passed to a linear classifiers), indep \\ (each local feature gets a linear classifier \\ of its own, the results of which are averaged).\end{tabular}} &  &  &  \\ \cline{1-2}
                                    &                                                                                                                                                                                                                                                                                                                                 &  &  & 
\end{tabular}
\caption{Hyperparameters of the Learn to Pay attention model}
\end{table}

\subsection{Dataset Preprocessing and Training Schedules}

We followed the preprocessing steps for CIFAR10, CIFAR100, and SVHN exactly as described by the authors of the paper both within the paper and within the OpenReview comments: ZCA Whitening was used on CIFAR10 and CIFAR100, and SVHN was not preprocessed. Code for preprocessing the datasets and packaging them is provided within our Github repository.

\section{Replication}

Table 2  reports the results of our replication. Top-1 errors for our replication trail just slightly behind for CIFAR-10 and CIFAR-100, except for SVHN where our replication performed better than the original.

\begin{table}[h]
\begin{tabular}{llll}
\hline
                                & CIFAR-10 & CIFAR-100 & SVHN \\ \hline
Original (VGG-att2)-concat-pc   & 5.23     & 23.9      &      \\
Replicated (VGG-att2)-concat-pc & 6.16     &      &      \\ \hline
Original (VGG-att3)-concat-pc   & 6.34     & 22.97     & 3.52 \\
Replicated (VGG-att3)-concat-pc &          & 24.11     & 3.21 \\ \hline
\end{tabular}
\caption{Top-1 classification errors}
\label{my-label}
\end{table}

\bibliographystyle{plain}
\bibliography{bib}

\end{document}